\pdfoutput=1

\documentclass[11pt]{article}

\usepackage[final]{acl}

\usepackage{times}
\usepackage{latexsym}
\usepackage{tikz}
\usepackage{amsfonts}
\usepackage{amsmath}
\usepackage{marvosym}
\usepackage{soul}

\usepackage[T1]{fontenc}

\usepackage{multirow}
\usepackage{xcolor}
\usepackage{listings}

\definecolor{codegreen}{rgb}{0,0.6,0}
\definecolor{codegray}{rgb}{0.5,0.5,0.5}
\definecolor{codepurple}{rgb}{0.58,0,0.82}
\definecolor{backcolour}{rgb}{0.95,0.95,0.92}

\lstdefinestyle{mystyle}{
    backgroundcolor=\color{backcolour},   
    commentstyle=\color{codegreen},
    keywordstyle=\color{magenta},
    numberstyle=\tiny\color{codegray},
    stringstyle=\color{codepurple},
    basicstyle=\ttfamily\footnotesize,
    breakatwhitespace=false,         
    breaklines=true,                 
    captionpos=b,                    
    keepspaces=true,                 
    numbers=left,                    
    numbersep=5pt,                  
    showspaces=false,                
    showstringspaces=false,
    showtabs=false,                  
    tabsize=2
}

\usepackage{microtype}

\usepackage{inconsolata}

\usepackage{graphicx}

\usepackage{booktabs}
\usepackage{latexsym}

\title{Explainable ICD Coding via Entity Linking\textit{}}

\author{
  Leonor Barreiros\textsuperscript{\Moon, \Libra, \Neptune} \quad
  Isabel Coutinho\textsuperscript{\Libra, \Neptune} \quad
  Gonçalo M. Correia\textsuperscript{\Moon} \quad
  Bruno Martins\textsuperscript{\Libra, \Neptune} \\
\textsuperscript{\Moon} Priberam Labs, Alameda D.\ Afonso Henriques, 41, 2º, 1000-123 Lisboa, Portugal\\
\textsuperscript{\Libra} Instituto Superior Técnico, Lisboa, Portugal \\
\textsuperscript{\Neptune} INESC-ID, Rua Alves Redol, 9, 1000-029, Lisboa, Portugal \\
\{\href{mailto:leonor.barreiros@priberam.pt}{\tt leonor.barreiros},
\href{mailto:goncalo.correia@priberam.pt}{\tt goncalo.correia}\}{\tt @priberam.pt}\\
\{\href{mailto:isabel.coutinho@tecnico.ulisboa.pt}{\tt isabel.coutinho}, \href{mailto:bruno.g.martins@tecnico.ulisboa.pt}{\tt bruno.g.martins}\}{\tt @tecnico.ulisboa.pt}
}

\begin{document}
\maketitle
\begin{abstract}
Clinical coding is a critical task in healthcare, although traditional methods for automating clinical coding may not provide sufficient explicit evidence for coders in production environments. This evidence is crucial, as medical coders have to make sure there exists at least one explicit passage in the input health record that justifies the attribution of a code. We therefore propose to reframe the task as an entity linking problem, in which each document is annotated with its set of codes and respective textual evidence, enabling better human-machine collaboration. By leveraging parameter-efficient fine-tuning of Large Language Models (LLMs), together with constrained decoding, we introduce three approaches to solve this problem that prove effective at disambiguating clinical mentions and that perform well in few-shot scenarios.
\end{abstract}

\section{Introduction}


Medical reports are essential documents that detail patient medical history, procedures, exams, symptoms, and diagnoses. Clinical coding involves assigning standardized codes, such as those from ICD-10, to these records. This process is crucial for hospitals, since it helps justify expenses, secure funding, or file insurance claims to cover healthcare costs. Furthermore, labeling Electronic Health Records (EHRs) through clinical coding makes their data more searchable and suitable for statistical analysis, \textit{e.g.} potentially revealing cause-effect relationships between diseases and symptoms.

Automated solutions can help medical coders by accelerating their work and reducing errors. However, traditional automated systems that treat coding as a Multi-Label Classification (MLC) problem are often non-explainable~\citep{icd_explain3, icd_explain4}, making it difficult for healthcare professionals to trust or verify their outputs. 
If systems are explainable, we can critically reason about their decisions, allowing medical practitioners to better work alongside AI tools~\citep{xai,responsible_ai}.

To address these challenges, we propose framing clinical coding as an entity linking problem. This particular task involves annotating documents with specific entities and providing textual evidence for each one. This could enable clinical coders to understand where each code is mentioned in a record, allowing easier cooperation with AI systems. 
However, clinical entity linking remains largely underexplored and lacking in terms of annotated data.

Recently, we have seen several advances in Transformer-based~\citep{transformer} Large Language Models (LLMs), such as LLaMA~\citep{llama-1}, Mistral~\citep{mistral}, or Gemini~\citep{gemini}, and in the formulation of data- and compute-efficient ways to fine-tune them~\citep{lora,qlora}. Consequently, we focus on mitigating the above challenges by exploring clinical entity linking as a generative task through a biomedical LLM, namely BioMistral~\citep{biomistral}. By fine-tuning an LLM, we aim to develop a system capable of solving clinical entity linking tasks effectively. 

Our contributions are three-fold: (i) we propose to frame the \textbf{explainability} of ICD coding as an entity linking task; (ii) we investigate the performance gains of prompting \textit{versus} fine-tuning a clinical LLM for this task, evaluating how different formulations for \textbf{generative entity linking} can enhance model performance; and (iii) we compare the entity linking approach to MLC, assessing the potential it has for \textbf{few-shot classification} of mentions.

\section{Proposed Approaches}

Traditionally, clinical coding is treated as MLC, in which a model annotates the input medical report with its set of labels.
In our setting, we treat clinical coding as an entity linking problem. This means that given a medical report and its set of gold mentions (\textit{i.e.}, our work assumes mentions have been pre-detected, for instance, via named entity recognition), our model must disambiguate each mention by assigning it the corresponding entity. 

The following subsections detail different approaches for tackling clinical entity linking.

\subsection{\textsc{ICL-BioMistral}}\label{generativeA}

\textsc{ICL-BioMistral} (in-context learning) prompts a pre-trained Transformer decoder model. 
The prompt comprises a (pre-determined) mention, and a medical report excerpt, corresponding to the context that surrounds it. The model must output an ICD-10 code representation, corresponding to the entity which the mention refers to.

Inspired by \citet{prompt-generative}, we designed a prompt with a short context and the task description. 
To improve the model's capability to solve the task, we use in-context learning (thoroughly analyzed by~\citet{icl}). 
As such, we add $10$ random examples to the prompt.
We illustrate the prompt template in Appendix~\ref{app:prompt}.

Similarly to \textsc{GENRE}~\citep{genre}, we use constrained greedy decoding,\footnote{\url{https://huggingface.co/blog/constrained-beam-search}} to ensure that the model output is always a valid ICD-10 code representation. 
This is implemented with a prefix tree of all possible outputs, and by forcing the generated tokens to stay within the set of possible continuations for titles of ICD codes.

\subsection{\textsc{SFT-BioMistral}}\label{generativeB}

\textsc{SFT-BioMistral} (supervised fine-tuning) is similiar to ICL-BioMistral, as it also outputs an in-context mention, given a report excerpt.
However, instead of learning through examples, this model was fine-tuned on a causal language modeling objective, where we maximize the conditional probability for each output token, considering the input and the expected previously generated output tokens~\citep{teacher_forcing}.
We consider as \textit{input} the prompt (\textit{i.e.}, the task description and context), and compute the cross-entropy loss over the tokens of the \textit{output} (the title of the desired ICD-10 code). Decoding with this model again relies on a constrained decoding algorithm.

\subsection{\textsc{InsGenEL-BioMistral}}\label{generativeC}

Our last proposed model is inspired by \textsc{InsGenEL}~\citep{insgenel}, which is based on \textsc{GENRE}~\citep{genre}.
Our model outputs multiple mention-entity pairs for a medical report in a single pass. 
This is closer to the approach clinical coders take when annotating, and it enriches predictions through the document's global context, improving coherence between predictions.

Like \textsc{GENRE}, our model receives a document (with gold mentions) and outputs the document with annotated mention-entity pairs. 
Unlike \textsc{GENRE}, and following \textsc{InsGenEL}, we use a Transformer decoder to annotate the documents.
The fine-tuning process optimizes a causal language modeling objective by learning from supervised instruction-response pairs~\citep{instruction_ft}. A prompt template is presented in Appendix~\ref{app:prompt}.

During inference, we ensure a valid generation using constrained decoding.
We implemented a function (based on \textsc{GENRE}'s proposal) that receives the generated tokens and returns the possible continuations.
First, it determines the state as either outside an entity---which can be the case when processing a non-mention or mention token---or inside an entity---where the model is disambiguating a mention. 
If outside an entity, then the possible continuation is to resume copying the input document. 
Otherwise, the model generates an entity representation.
Similarly to our previous approaches, we use a prefix tree to ensure the model generates valid ICD-10 code representations. 

\section{Experimental Setup}

To train and evaluate our models, we used publicly available datasets for explainable ICD coding, \textit{i.e.} including span evidences for each code, namely CodiEsp~\citep{codiesp}, DisTEMIST~\citep{distemist}, and MDACE~\citep{mdace}. Further details on these datasets are given in Appendix~\ref{app:datasets}.
Additional experimental details are given in Appendix~\ref{app:details}.

\paragraph{Knowledge Base.}
In entity linking, entities are organized in knowledge bases. 
We focus on the International Classification of Diseases (ICD)\footnote{\url{https://www.who.int/standards/classifications/classification-of-diseases}} coding system, proposed by the World Health Organization, as a standardized way of representing diagnoses and procedures.
The ICD is a hierarchical ontology, as codes are first organized into chapters, sub-chapters, and partial codes. 
We considered version 10, which is divided into ICD-10-CM (for diagnoses) and ICD-10-PCS (for procedures).

\paragraph{Evaluation Details.}\label{evaluation_details}
In end-to-end entity linking, we distinguish the precision, recall, and F1 metris. In our case, where we used gold mentions, these equate to a measure of accuracy, as explained by~\citet{entity_linking_chapter}. We consider micro-accuracy (where we average the accuracy of all mentions) and macro-accuracy (where we compute the accuracy per document and average all values). 
To compare our results with existing work, we computed coding evaluation metrics. By aggregating all assignments for the entity linking task, we obtain a solution for MLC that can be evaluated with precision, recall, and F1. These were computed with the script by \citet{codiesp}.

\section{Experimental Results}

Table~\ref{tbl:accuracy} presents our micro- and macro-accuracy on the CodiEsp and MDACE test datasets.

\begin{table}[t!]
\centering
\begin{tabular}{llrr}\toprule
& & Micro & Macro \\ \midrule
\multirow{3}{*}{\rotatebox[origin=c]{90}{CodiEsp}} & \textsc{ICL-BM} & $6.36$ & $5.93$ \\ 
& \textsc{SFT-BM} & $63.39$ & $62.41$ \\ 
& \textsc{InsGenEL-BM} & $\mathbf{66.85}$ & $\mathbf{64.40}$ \\ \midrule
\multirow{3}{*}{\rotatebox[origin=c]{90}{MDACE}} & \textsc{ICL-BM} & $10.36$ & $7.79$ \\
& \textsc{SFT-BM} & $\mathbf{64.88}$ & $\mathbf{60.94}$ \\
& \textsc{InsGenEL-BM} & $57.10$ & $55.45$ \\
\bottomrule
\end{tabular}
\caption{Accuracy in the CodiEsp and MDACE test sets for the entity linking task. \textsc{BM} denotes \textsc{BioMistral}. We highlight in bold the best-in-class performance.}
\label{tbl:accuracy}
\end{table}

\paragraph{Practical Highlights.}\label{p:gen_usability}
From Table~\ref{tbl:accuracy}, we conclude that fine-tuned models perform considerably better than \textsc{ICL-BioMistral}. We highlight that \textsc{SFT-BioMistral} has a stable performance for both evaluation corpora, whereas \textsc{InsGenEL-BioMistral} has limitations in MDACE, which
we hypothesize might be related to the increased length of the documents.
Additionally, we find that \textsc{InsGenEL-BioMistral} is beneficial in production scenarios: not only does it better alleviate the coder's job with its increased accuracy, but it also deals with all of a document's mentions simultaneously. 
Nonetheless, clinical coders receive non-annotated documents and a separate procedure must be used to recognize and annotate the textual evidence to which a code should be assigned.

\paragraph{Partial Results.} Since the ICD-10 is organized hierarchically, a wrong prediction can be partially correct if it determines the code's ancestors up to a certain point. 
We assessed micro-accuracy on the chapter, subchapter, and partial code levels (a partial code contains the first three digits of an ICD), and the results are in Table~\ref{tbl:partial_results}.
Both \textsc{SFT-BioMistral} and \textsc{InsGenEL-BioMistral} can provide orientation helpful in practical scenarios. 

\begin{table}[t!]
\centering
\begin{tabular}{llrrr}\toprule
& & \multicolumn{1}{r}{Chap} & \multicolumn{1}{r}{Sub} & \multicolumn{1}{r}{Part} \\
\midrule
\multirow{3}{*}{\rotatebox[origin=c]{90}{CodiEsp}} & \textsc{ICL-BM} & $30.64$ & $18.33$ & $12.55$ \\ 
& \textsc{SFT-BM} & $85.65$ & $82.27$ & $75.94$ \\ 
& \textsc{InsGenEL-BM} & ${87.79}$ & ${83.60}$ & ${78.81}$ \\ \midrule
\multirow{3}{*}{\rotatebox[origin=c]{90}{MDACE}} & \textsc{ICL-BM} & $43.88$ & $33.90$ & $23.35$ \\ 
& \textsc{SFT-BM}      & $89.17$ & ${84.84}$ & ${78.91}$ \\ 
& \textsc{InsGenEL-BM} & ${90.09}$ & $83.73$ & $76.18$ \\ 
\bottomrule
\end{tabular}
\caption{Micro-accuracy in the CodiEsp and MDACE test sets for the entity linking task, considering only the chapter (Chap), subchapter (Sub), and partial (Part) code of each ICD-10. \textsc{BM} denotes \textsc{BioMistral}.}
\label{tbl:partial_results}
\end{table}

\begin{table}[t!]
\centering
\begin{tabular}{llrr}\toprule
& & \multicolumn{1}{r}{$1$-shot} & \multicolumn{1}{r}{$5$-shot} \\ \midrule
\multirow{2}{*}{CodiEsp} & \textsc{SFT-BM} & $\mathbf{47.49}$ & $\mathbf{56.66}$  \\
& \textsc{InsGenEL-BM} & $34.97$ & $49.30$ \\ \midrule
\multirow{2}{*}{MDACE} & \textsc{SFT-BM} & $\mathbf{36.74}$ & $\mathbf{40.89}$  \\
& \textsc{InsGenEL-BM} & $24.39$ & $29.66$ \\
\bottomrule
\end{tabular}
\caption{$1$- and $5$-shot micro-accuracy in the CodiEsp and MDACE test corpora. \textsc{BM} denotes \textsc{BioMistral}.}
\label{tbl:1-shot-results}
\end{table}

\begin{table*}[t!]
\centering
\begin{tabular}{lrrrrrrrrr}\toprule
& \multicolumn{6}{r}{Multi-label Classification} & \multicolumn{3}{r}{Entity Linking} \\ \cmidrule(lr){2-7} \cmidrule(lr){8-10}
& \multicolumn{3}{r}{CodiEsp-D} & \multicolumn{3}{r}{CodiEsp-P}  & \multicolumn{3}{r}{CodiEsp-X} \\ \cmidrule(lr){2-4} \cmidrule(lr){5-7} \cmidrule(lr){8-10}
& \multicolumn{1}{r}{P} & \multicolumn{1}{r}{R} & \multicolumn{1}{r}{F1} & \multicolumn{1}{r}{P} & \multicolumn{1}{r}{R} & \multicolumn{1}{r}{F1} & \multicolumn{1}{r}{P} & \multicolumn{1}{r}{R} & \multicolumn{1}{r}{F1} \\
\midrule
IAM CodiEsp & $\mathbf{81.70}$ & $59.20$ & $68.70$ & $\mathbf{69.10}$ & $42.00$ & $52.20$ & $\mathbf{75.00}$ & $52.40$ & $61.10$\\ 
DAC-E & $-$ & $-$ & $74.40$ & $-$ & $-$ & $\mathbf{56.0}$ & $-$ & $-$ & $-$ \\
\midrule
\textsc{ICL-BM} & $8.91$ & $7.19$ & $7.96$ & $11.34$ & $12.45$ & $11.87$ & $8.42$ & $7.19$ & $7.76$ \\ 
\textsc{SFT-BM} & $75.04$ & $\mathbf{76.20}$ & $\mathbf{75.62}$ & $34.31$ & $38.53$ & $36.30$ & $64.66$ & $\mathbf{67.10}$ & $65.86$ \\ 
\textsc{InsGenEL-BM} & $73.93$ & $71.94$ & $72.92$ & $46.26$ & $\mathbf{46.78}$ & $46.52$ & $68.34$ & $66.96$ & $\mathbf{67.64}$ \\ 
\bottomrule
\end{tabular}
\caption{Comparison of automated medical coding and entity linking micro performance metrics on the CodiEsp test set with existing results for the CodiEsp shared task. \textsc{BM} denotes \textsc{BioMistral}.}
\label{tbl:comparison_generative_codiesp}
\end{table*}

\paragraph{Few-shot Analysis.}\label{p:few-shot}
In Table~\ref{tbl:1-shot-results}, we compare the few-shot performance for all codes seen at most once or $5$-times during training ($1$-shot and $5$-shot). The number of such codes in the inference corpora is given in Appendix~\ref{app:datasets}.
The model with the best few-shot performance was \textsc{SFT-BioMistral}, but \textsc{InsGenEL-BioMistral} is nevertheless able to predict codes trained in few-shot scenarios. We hypothesize that the reduced performance on MDACE is related to the increased document length, which may lead to hard long-range dependencies.

\subsection{Comparison with Existing Results} 
 
The CodiEsp-D and CodiEsp-P tasks can be evaluated with MLC metrics, as we explain in \S\ref{evaluation_details}.
CodiEsp also proposes an end-to-end entity linking task, CodiEsp-X. It is not evaluated with entity linking metrics, since if a code is mentioned more than once in the same document, it only needs to be correctly predicted once to be considered correct. This means the evaluation micro-metrics for CodiEsp-X do not equate to our micro-accuracy.

In Table~\ref{tbl:comparison_generative_codiesp}, we compare our results to those of the challenge's winner, \textit{i.e.}, the IAM team~\citep{iam_codiesp}, and to a solution that was subsequently proposed, DAC-E~\citep{codiesp_outro}. These systems are described in Appendix \ref{app:baselines}. 
Although a strict comparison is not possible, since we used gold mentions contrarily to the shared tasks, our fine-tuned models had similar or better performance in most settings, indicating that our approaches remain useful in the MLC scenario.

MDACE was proposed for a different task: given the output of MLC, finding sufficient textual evidence for each code.
This means that we cannot compare with the paper's benchmarking results.

\section{Related Work}

We briefly describe previous related work on automated ICD coding and also on entity linking.

\paragraph{ICD Coding \& Explainability.} Most solutions for automated ICD coding are based on MLC. 
For example, \citet{codiesp_outro} leverage the ICD hierarchy and propose two MLC sub-tasks on different granularities.
Furthermore, many studies have addressed the importance of solving explainable ICD coding, so that clinical coders can understand the system's decisions. 
However, most studies focus on label-wise attention mechanisms~\citep{icd_explain,icd_explain2,priberam}, which are challenging to systematically evaluate, as pointed out by~\citet{icd_explain3} and~\citet{icd_explain4}. 
More recently, researchers have developed methodologies to evaluate these interpretability solutions~\citep{icd_explain5, icd_explain6}.

\paragraph{Entity Linking \& Different Entity Linking Approaches.}
Entity linking solutions range from discriminative to generative models. 
Discriminative models are the most common, but many state-of-the-art models, such as those of~\citet{luke},~\citet{refined}, and~\citet{spel}, were trained on large corpora (the Wikipedia), which is not available for our domain.
Generative models require less fine-tuning data to achieve similar performance. For example, \citet{insgenel} performed better than \citet{refined}, using $50$ times less data.
The model was inspired by a previous proposal from~\citet{genre}, which uses constrained decoding to ensure valid generation.

\paragraph{Clinical \& Biomedical Entity Linking.}
The clinical and biomedical domains are specialized, and general-purpose models cannot solve clinical problems, even with a target domain fine-tuning corpus~\citep{biomedical_el}.
Existing work uses methodologies similar to general-domain algorithms, but with models trained on domain corpora~\citep{biobart,biobart3}. For instance,~\citet{biobart2} propose a method similar to GENRE.
In the clinical domain, most entity linking studies focus on the DisTEMIST~\citep{distemist} and CodiEsp~\citep{codiesp} challenges. For example, \citet{clinlinker} propose a Transformer encoder-based solution to DisTEMIST. 

\section{Conclusions}

We described three approaches for the clinical entity linking problem, based on BioMistral 7B, that annotate medical reports with each mention's ICD-10 code. The models we fine-tuned, \textit{i.e.}, \textsc{SFT-} and \textsc{InsGenEL-BioMistral}, were substantially better than the prompted \textsc{ICL-BioMistral}, and yielded interesting results for few-shot codes. 

\section*{Limitations}
Our models only deal with the disambiguation sub-problem of entity linking, using pre-detected mentions. Future work 
should explore mention detection to obtain an end-to-end solution, which makes our models useful in production environments. 

In addition, our experiments were limited to three publicly available datasets, which only represent a small subset of patients, possible medical conditions, and medical procedures.
There is not a lot of clinical data publicly available to support research studies, especially annotated for entity linking. In the future, we can explore other approaches to data collection, and even leverage additional information from clinical knowledge bases, such additional information in ICD-10 itself and UMLS.

Finally, large generative models such as BioMistral 7B are generally very costly to use. For instance, the IAM system~\citep{iam_codiesp}, based on a dictionary, only takes 5 seconds to run on an 8 CPUs' machine. The DAC-E~\citep{codiesp_outro} system, while using GPU processing, is also more efficient as it uses a smaller Transformer encoder as the backbone. Future work can perhaps assess the impact of using LLMs of different sizes.

\section*{Ethical Considerations}

ICD coding is a sensitive task that influences clinical and financial decisions. In our problem formulation, we facilitate keeping practitioners in charge of all clinical decisions, as they can critically assess each model decision. 
This allows medical coders to work alongside AI tools, fostering human-machine collaboration rather than replacing human input, with basis on the supporting evidence. 

Due to restrictions in data access, we used publicly available datasets that only represent a small part of the target population.
To use the MDACE corpus, we took the \textit{Data or Specimens Only Research} training course from the CITI program.\footnote{\url{https://about.citiprogram.org/}}

\section*{Acknowledgments}

This research was supported by the Portuguese Recovery and Resilience Plan through project C645008882-00000055 (i.e., the Center For Responsible AI).

\bibliography{custom}

\appendix
\section{Prompt Templates}
\label{app:prompt}

The prompt used for \textsc{ICL-BioMistral} is in Listing~\ref{code:prompt_generativeA}.
For \textsc{SFT-BioMistral}, we used a similar prompt, without the \texttt{[Example]}s.
For \textsc{InsGenEL-BioMistral}, we used the prompt in Listing~\ref{code:prompt_generativeC}. We use a prompt in English, and generate outputs in English, even with CodiEsp's Spanish reports.

\section{Dataset Details and Statistics}
\label{app:datasets}

\begin{table*}[t!]
\centering
\begin{tabular}{lrrrrrrr} \toprule
& & \multicolumn{3}{r}{Diagnoses} & \multicolumn{3}{r}{Procedures} \\ \cmidrule(lr){3-5} \cmidrule(lr){6-8}
& {Reports} & {Samples} & {Codes} & {1-shot codes}  & {Samples} & {Codes} & {1-shot codes}\\ \midrule
CodiEsp   & $500$ & $8,199$ & $1,720$ & $618$ & $2,799$ & $435$ & $86$ \\
DisTEMIST & $750$ & $1,912$ & $451$ & $176$ & $23$ & $4$ & $1$ \\
MDACE     & $181$ & $4,993$ & $966$ & $446$ & $168$ & $89$ & $61$\\ \midrule
Total   & $1,431$ & $15,104$ & $2,513$ & $912$ & $2,990$ & $515$ & $138$ \\ \bottomrule
\end{tabular}
\caption{Datasets used for training. \textit{Codes} refers to the number of distinct ICD-10 codes in the training data, and \textit{1-shot codes} refers to the number of codes that only appear once.}
\label{tbl:datasets_train}
\end{table*}

\begin{figure}[t]
\begin{lstlisting}[caption={Prompt for \textsc{ICL-BioMistral}.}, label={code:prompt_generativeA}]
You are a medical coder at a hospital, and you have to assign ICD-10 codes to mentions. I will give you a report excerpt and a mention that can be found in that excerpt. Your job is to associate the mention to an ICD-10 code. 
Each code can be a Diagnosis in ICD-10-CM or a Procedure in ICD-10-PCS. You should give the ICD-10 code according to its type (Diagnosis or Procedure).
[Example]:
The following report excerpt, written in <language>: """<example_mention_in_context>""", contains the following mention: <example_mention>.
It corresponds to the ICD-10 entity: <example_icd>.
[Task]:
The following report excerpt, written in <language>: """<mention_in_context>""", contains the following mention: <mention>.
It corresponds to the ICD-10 entity:
\end{lstlisting}
\end{figure}

\begin{table}[t]
\centering
\begin{tabular}{lrr}\toprule
& \multicolumn{1}{r}{CodiEsp} & \multicolumn{1}{r}{MDACE} \\ \midrule
No. $1$-shot codes & $219$ & $49$ \\
No. $5$-shot codes & $923$ & $203$ \\
\bottomrule
\end{tabular}
\caption{Number of $1$-shot and $5$-shot codes in the CodiEsp and MDACE test sets, considering the number of times they were seen in the training corpus.}
\label{tbl:few-shot-stats}
\end{table}

\begin{figure}[t!]
\begin{lstlisting}[caption={Prompt for \textsc{InsGenEL-BioMistral}.}, label={code:prompt_generativeC}]
You are a medical coder at a hospital, and you have to assign ICD-10 codes to mentions.
I will give you a medical report, whose mentions are annotated between { and }. Your job is to associate each mention to an ICD-10 code. 
Each code can be a Diagnosis in ICD-10-CM or a Procedure in ICD-10-PCS. You should give the ICD-10 code according to its type (Diagnosis or Procedure) and hierarchy, that is, you should first write  the chapter, then the subchapter up until the title of the ICD-10 code, separated by "-->".
ICD-10 codes should be delimited by | and |.
Annotate the following report:
<report>
\end{lstlisting}
\end{figure}

We used three different corpora during training.

\begin{itemize}
    \item CodiEsp~\citep{codiesp} consists of Spanish medical reports, which were manually annotated with their ICD-10 codes and textual evidence spans. The corpus was developed for the CodiEsp shared task, which comprises three sub-tasks: automated ICD coding for ICD-10-CM (CodiEsp-D) and ICD-10-PCS (CodiEsp-P), and end-to-end clinical entity linking for ICD-10 (CodiEsp-X).
    \item DisTEMIST~\citep{distemist} comprises medical reports in Spanish and English (we only used the English version), manually annotated with their SNOMED CT disease codes and textual evidence spans. 
    The authors mapped the SNOMED CT codes to ICD-10 using UMLS. This mapping was only performed for the training data, so we could not evaluate our model's performance on the DisTEMIST validation and test splits.
    \item MDACE~\citep{mdace} consists of English medical reports, which are part of the MIMIC-III collection~\citep{mimic_iii}, with manually annotated ICD-10 codes and respective textual evidence spans.
\end{itemize}

The number of test few-shot codes is in Table~\ref{tbl:few-shot-stats}.
Table~\ref{tbl:datasets_train} summarizes the training datasets. 

\section{Experimental Details}\label{app:details}

Our models were initialized with BioMistral-7B~\citep{biomistral}. \textsc{SFT-} and \textsc{InsGenEL-BioMistral} were fine-tuned for $5$ epochs on an NVIDIA RTX A6000 GPU for $20$ hours, with a batch size of $4$. We used QLoRA~\citep{qlora}, with rank $r = 64$ and $4$-bit NF quantization, and the AdamW~\citep{adamW} optimizer with a learning rate of $2*10^{-4}$ and weight decay equal to $10^{-3}$.
For inference, models were loaded without quantization on the same GPU, and we used the same batch sizes and a greedy decoding strategy. Inference took $8$ hours for all datasets.

For \textsc{InsGenEL-BioMistral}, to ensure all training samples did not exceed the model's context window of $8,192$ tokens, we truncated all documents to $5,000$ characters.
During inference, the entire documents were processed.

\section{Comparison Systems}\label{app:baselines}

In Table \ref{tbl:comparison_generative_codiesp}, we compare our experimental results on the CodiEsp test corpus with those of the IAM and DAC-E systems, which work as follows:
\begin{itemize}
    \item The IAM~\citep{iam_codiesp} system performs explainable ICD coding. It starts by normalizing every document in the training data, and composing a dictionary whose items are the normalized mentions (denoted \textit{terms}) and their corresponding ground-truth ICD-10 codes. 
    Additionally, the KB entities' normalized titles are added to the dictionary. 
    Then, each dictionary term is tokenized and stored in an $n$-gram tree. 
    For inference, a matching algorithm parses each document's tokens to find matching dictionary entries. Three matching strategies are employed: perfect matching, abbreviation matching (where a hand-crafted dictionary of abbreviations is used), and Levenshtein distance-based matching.
    
    \item The DAC-E~\citep{codiesp_outro} approach is not as directly explainable, as it treats ICD coding as MLC. This system comprises two sub-tasks, respectively performed by \textit{matcher} and \textit{ranker} models. The matcher associates documents to clusters (the chapters in ICD-10), leveraging a biomedical RoBERTa model~\citep{roberta}. The ranker computes the likelihood of each code being present in a document, considering its chapter. It was implemented with a binary classifier for each code, trained only with documents with codes in the same cluster, for better fine-grained differentiation. The ranker was trained using the XGBoost algorithm~\citep{xgboost}.
\end{itemize}

\end{document}